%% file: main.tex
\documentclass[10pt,twocolumn,letterpaper]{article}

\usepackage{cvpr}              % To produce the CAMERA-READY version
\usepackage{graphicx}
\usepackage{amsmath}
\usepackage{amssymb}
\usepackage{booktabs}
\usepackage{epsfig}
\usepackage{multirow}
\usepackage[accsupp]{axessibility}

\usepackage{url}
\usepackage{epsfig}
\usepackage{graphicx}
\usepackage{comment}
\usepackage{adjustbox}
\usepackage{subcaption}
\usepackage{tabularx}
\usepackage[noend]{algpseudocode}
\usepackage[ruled]{algorithm2e}
\usepackage{flushend}

\usepackage[pagebackref,breaklinks,colorlinks]{hyperref}

\usepackage[capitalize]{cleveref}
\crefname{section}{Sec.}{Secs.}
\Crefname{section}{Section}{Sections}
\Crefname{table}{Table}{Tables}
\crefname{table}{Tab.}{Tabs.}

\makeatletter 
\g@addto@macro{\@algocf@init}{\SetKwInOut{Parameter}{Parameters}} 
\makeatother

\def\cvprPaperID{13} % *** Enter the CVPR Paper ID here
\def\confName{EarthVision}
\def\confYear{2023}

\input{math_commands.tex}

\begin{document}

\title{Cascaded Zoom-in Detector for High Resolution Aerial Images}
%First Author\\
\author{Akhil Meethal
\and Eric Granger
\and Marco Pedersoli \and \vspace{-1em}\\
{LIVIA lab,  Dept. of Systems Engineering, ÉTS Montreal, Canada} \\
{\tt\small akhilpm135@gmail.com}, 
{\tt\small \{marco.pedersoli, eric.granger\}@etsmtl.ca}
}
\maketitle

\input{0-abstract}

\input{1-introduction}
\input{2-related-work}
\input{3-proposal}
\input{4-experiments}

\input{5-conclusion}
%\input{6-supp-materials}

%%%%%%%%% REFERENCES
{\small
\bibliographystyle{ieee_fullname}
\bibliography{egbib}
}

\input{6-supp-materials}

\end{document}

%% file: math_commands.tex
\usepackage{amsmath,amsfonts,bm}

\def\eqref#1{equation~\ref{#1}}
\def\1{\bm{1}}

\DeclareMathAlphabet{\mathsfit}{\encodingdefault}{\sfdefault}{m}{sl}
\SetMathAlphabet{\mathsfit}{bold}{\encodingdefault}{\sfdefault}{bx}{n}

\DeclareMathOperator*{\argmax}{arg\,max}

%% file: 0-abstract.tex
% !TEX root=main.tex

\begin{abstract}
Detecting objects in aerial images is challenging because they are typically composed of crowded small objects distributed non-uniformly over high-resolution images. Density cropping is a widely used method to improve this small object detection where the crowded small object regions are extracted and processed in high resolution. However, this is typically accomplished by adding other learnable components, thus complicating the training and inference over a standard detection process. In this paper, we propose an efficient Cascaded Zoom-in (CZ) detector that re-purposes the detector itself for density-guided training and inference. During training, density crops are located, labeled as a new class, and employed to augment the training dataset. During inference, the density crops are first detected along with the base class objects, and then input for a second stage of inference. This approach is easily integrated into any detector, and creates no significant change in the standard detection process, like the uniform cropping approach popular in aerial image detection. Experimental results on the aerial images of the challenging VisDrone and DOTA datasets verify the benefits of the proposed approach. The proposed CZ detector also provides state-of-the-art results over  uniform cropping and other density cropping methods on the VisDrone dataset, increasing the detection mAP of small objects by more than 3 points. \footnote{Our code is available at: \href{url}{https://github.com/akhilpm/DroneDetectron2}}
\end{abstract}

%% file: 1-introduction.tex
% !TEX root=main.tex

\section{Introduction}
\label{sec:introduction}
\begin{figure}[t]
  \centering
  \includegraphics[height=6.7cm, width=0.49\textwidth]{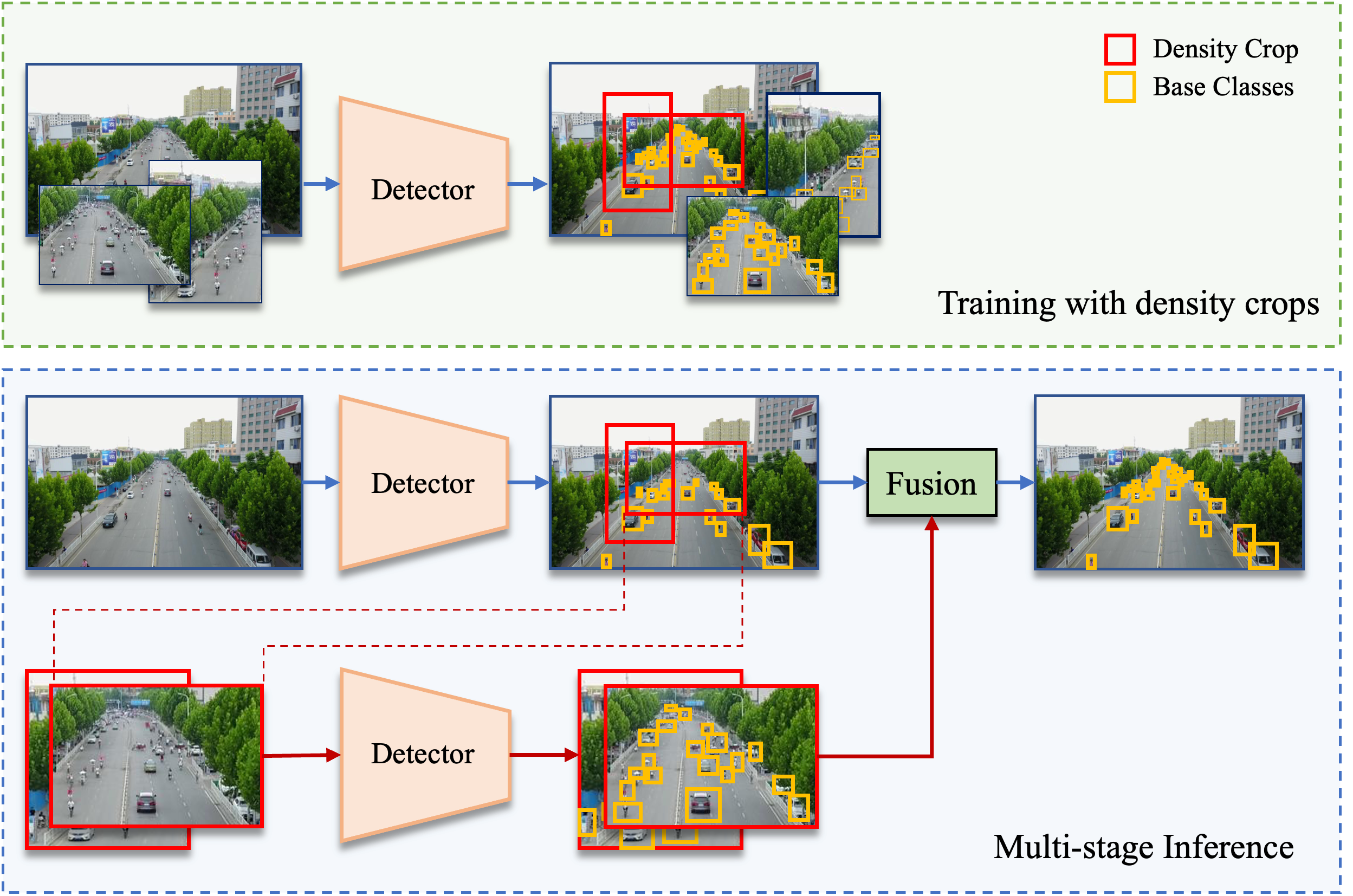} 
  \caption{Overview of our proposed Cascaded Zoom-in detector. During training (top), density crops are extracted, and labeled as a new class (red boxes) on the original image. The training set is augmented with the rescaled density crops and the corresponding ground truth boxes within these crops. During the first stage of inference (bottom), the base class objects and density crops (red boxes) are detected on the whole image. In the second stage, the density crops are rescaled to a common larger size, and a second inference is performed. Finally, the detections on density crops are combined with the detections on the whole image.}
  \label{fig:base_daigram}
\end{figure}

With the advent of deep learning, object detection methods achieved significant progress \cite{fpn-Lin-2017, detr-Carion-2020, cascadercnn-Cai-2018, fcos-Tian-2019, faster_rcnn-Ren-2015, yolo_9000-Redmon-2017}. This has resulted in the fast growth of their adoption to many downstream applications, including aerial images captured with Drones or Satellites, for earth monitoring, surveillance, inspection, etc \cite{geobench-Alex-2021, xia-DOTA-2018, aerial1-Cheng-2016, aerial2-Han-2021, remote_sense-Long-2017}. However, unlike natural images in the Pascal VOC \cite{voc-Everingham-2010} and MS-COCO \cite{mscoco-Lin-2014} datasets, aerial images are captured in high resolution, and are typically comprised of many small objects, that are sparsely distributed in crowded object regions. As a comparison, the average number of objects in Pascal VOC and MS-COCO images are 3 and 7, respectively, whereas images in the VisDrone \cite{zhu-VisDrone-2018} and DOTA \cite{xia-DOTA-2018} datasets -- two popular benchmarks in the aerial detection community -- have an average number of 53 and 67 objects, respectively. The average width of Pascal VOC and MS-COCO images are 500 and 640 pixels, respectively, while  the same in VisDrone and DOTA images are 1500 and 4000 pixels, respectively. Therefore, improvements observed in object detection methods applied to natural images do not easily translate to object detection in high-resolution aerial images.

\begin{figure*}[h!]
  %\centering
  \hspace{1cm}
  \includegraphics[height=7.4cm, width=0.9\textwidth]{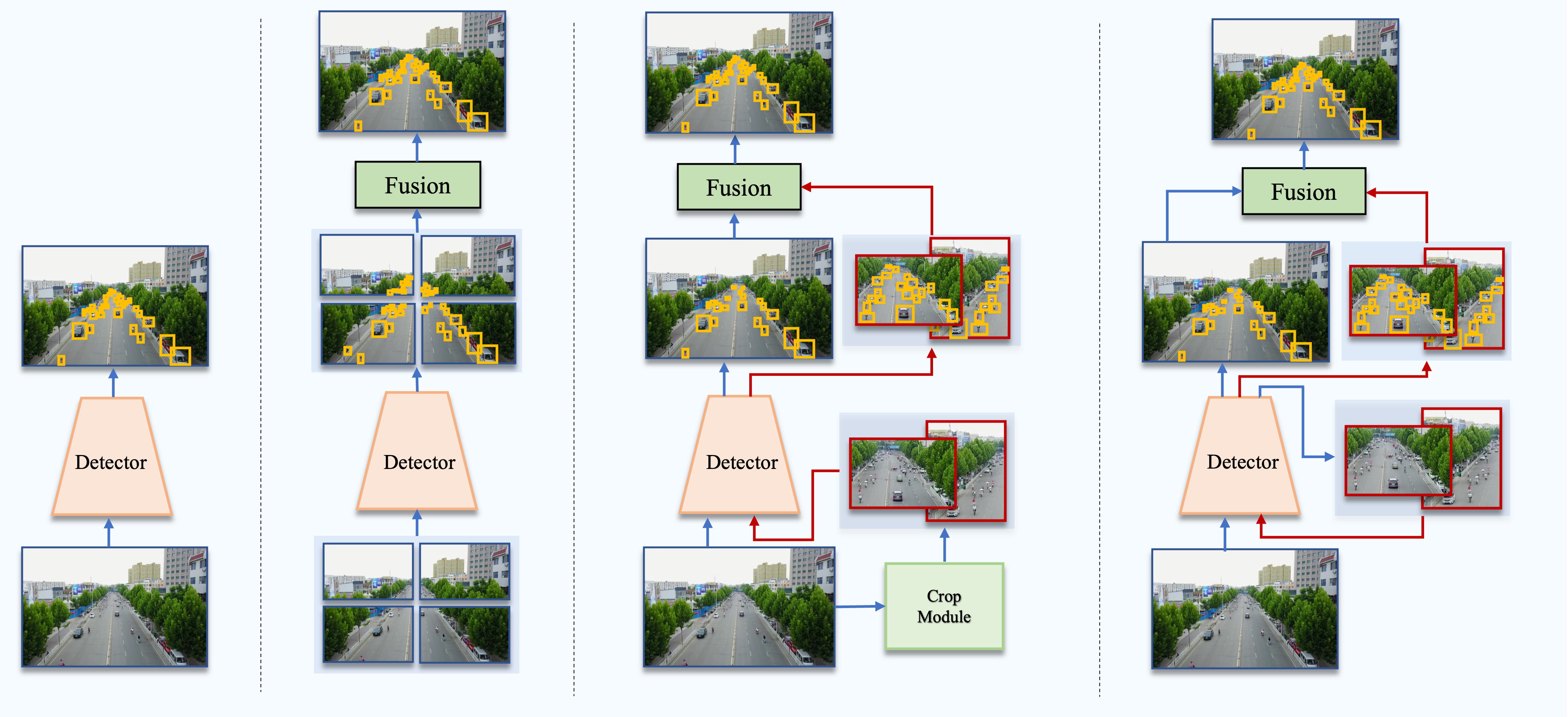} \\
  %\hspace{1cm}
  \begin{tabular}{@{\hspace{+1.3cm}} l @{\hspace{+1.1cm}} l  @{\hspace{1.3cm}} l @{\hspace{+2.3cm}} l}%{cccc}
     (a) Baseline & (b) Uniform crops & (c) Density crops & (d) Our approach \\
  \end{tabular}
  \caption{An illustration of different approaches for the detection of small objects in high-resolution aerial images. (a) The image is down-scaled and processed at the detector's resolution. (b) The image is split into uniform, possibly overlapping crops, and each crop is processed by the detector. (c) An external learnable module crops the image into dense object regions. Each crop is re-scaled and processed at the detector's resolution. (d) Our proposed CZ detector is re-purposed to detect the density crops along with the base class objects, eliminating the need for an external module. Each crop is re-scaled and processed at the detector's resolution in a second stage of inference. Blue arrows show the path of the original image and red shows the path of density crops.}
  \label{fig:main_diagram}
\end{figure*}

The high-resolution imagery and tiny objects raise several challenges for object detection in aerial images, including the loss of information due to rescaling and feature down-sampling, low tolerance to bounding box shifts, and noisy feature representations \cite{sod_survey-Cheng-2022, glsan-Deng-2020} among others. Since it is difficult to input high-resolution aerial images directly to a detector due to the computational cost and large memory footprint, they are often resized to the standard resolution range (see Fig. \ref{fig:main_diagram}~(a)). This rescaling, coupled with the feature down-sampling in ConvNets, often results in feature representations linked to small objects diminished or corrupted by the noisy background activations \cite{querydet-Yang-2022}. Regarding the tolerance to bounding box shifts, a small shift in the bounding box may cause a large decline in the Intersection over Union (IoU), raising the false positive detections \cite{sod_survey-Cheng-2022}. 

To utilize the higher resolution, and also mitigate the information loss, a popular approach consists in cropping the input image into uniform patches, and then performing object detection on these patches in high resolution by upsampling (see Fig. \ref{fig:main_diagram}~(b)). Although these uniform patches help to improve the accuracy, this approach does not respect the distribution of the objects in the image, and hence the scale normalization achieved is not optimal \cite{clusnet-Yang-2019, dmap-Li-2020}. As the objects in aerial images usually appear crowded in sparsely distributed regions of the image, it is desirable to perform density-based cropping, and then process high-resolution versions of those crowded object regions for better scale balance. 

To extract these density crops, existing methods utilizes additional learnable modules (see Fig. \ref{fig:main_diagram}~(c)) such as density maps \cite{dmap-Li-2020}, cluster proposal networks \cite{clusnet-Yang-2019}, global-local detection pipeline \cite{glsan-Deng-2020} etc. This usually results in additional learnable components in the pipeline, often with multiple stages of training. Even with single-stage end-to-end methods, the crops obtained are noisy in the beginning and are only useful for aiding small object detection in the later stages of the training \cite{clusnet-Yang-2019}. Moreover, as the learning is evolving the crops also evolve, so the model is not receiving a consistent indication of what exactly is a "density crop". Thus, practitioners still widely use uniform cropping over the advanced density crop-based approaches due to the practical simplicity it offers \cite{small_obj_detection, tiling_obj_detection}.

In this paper, we tried to bridge this gap between research and practice by proposing an efficient cascaded zoom-in (CZ) detection method that can leverage density crops within the training of a standard detector, offering the simplicity of uniform cropping, yet providing the benefits of density crops(see Fig. \ref{fig:main_diagram}~(d)). We simply make use of the existing detector itself to discover the density crops, by adding the "crop" as a new class to the detector. The crops are labeled as a pre-processing step using a crop labeling algorithm, and hence the detector receives a consistent signal of what constitutes a crop. During inference, while other methods require complex post-processing to filter the noisy crops, we can simply perform it based on the confidence of the "crop" class from the detector.

Fig.~\ref{fig:base_daigram} illustrates the training and testing of our method. First, the density crops are extracted from each training image as a pre-processing step, using our crop labeling algorithm. These density crops are added as a new class to be detected in the corresponding image. Then we augment the training set with the higher resolution version of the density crops, and the corresponding ground truth (GT) boxes of objects inside the crop. Then, the detector is trained as usual. This training process has an almost negligible overhead over standard detector training, and it is similar to that of uniform crop based training. Inference is performed in two stages. In the first stage, the base class objects and density crops are detected from each input image. In the second stage, high-quality density crops are selected based on their confidence score, and another inference based is performed on an up-sampled version of these crops. Finally, the detections from stages one and two are fused to get the output detection. Compared to standard object detector learning, the extra work required at training time is the crop labeling which can be performed as a pre-processing step. While making predictions, the extra work required is one more inference. As both of these processes don't require any significant modification on a normal object detection pipeline, similar to uniform crops, our method can be easily incorporated for accelerating small object detection.

Our main contribution can be summarized as follows:

\noindent \textbf{(1)} An efficient cascaded zoom-in (CZ) object detector based on density cropping is proposed for high-resolution aerial images, where a given detector is re-purposed to extract density crops(thus not using additional modules) along with base class objects. At train time, this approach relies on a simple pre-processing step for density crop labeling, and at test time, an additional inference step is needed to process the predicted density crops.

\noindent \textbf{(2)} We empirically validate the benefits of our cascaded zoom-in detection on aerial images from drones (VisDrone) and satellites (DOTA), and obtain state-of-the-art performance. A consistent improvement in the detection accuracy is observed on both datasets over the baseline detection. Particularly, the mAP of small object detection is improved by more than 3 points.

%In contrast, our method proposes to use the detector not only to find the classes of interest, but also to accurately localize those regions in the image containing groups of objects. This reduces the complexity of the approach as only an "augmented" detector needs to be trained.  and allows us to use any detector for this approach.
%We initially use the detector at coarse resolution and zoom-in only on the regions that are classified as density crops and need to be further analyzed at higher resolution. This allows a dynamic analysis of the image, in which only the regions with a high probability to contain objects are analyzed at high resolution, reducing the computational time of the method and also possible false positives that would be generated by densely analyzing the image at high resolution.
%In addition to that, they often used overly complex learning algorithms to train those learnable components often as a separate stage from the detector training. In this paper, we propose that scale normalization can be simply achieved by re-purposing the detector with a multi-stage inference. So without any additional learnable components and learning algorithms, we detect the focus regions to adaptively crop and zoom into 

%  \includegraphics[scale=0.57]{images/base_diagram.png} 
%  \includegraphics[height=4cm, width=0.23\textwidth]{images/base_diagram.png} 

%% file: 2-related-work.tex
% !TEX root=main.tex

\section{Related works}
\label{sec:related-work}

%\subsection{Object Detection}
\noindent \textbf{Object Detection.}  
General purpose visual object detectors are built primarily focusing on natural images. They can be broadly classified into two-stage \cite{faster_rcnn-Ren-2015, fpn-Lin-2017} and one-stage \cite{yolo_9000-Redmon-2017, ssd-Liu-2016} object detectors. Two-stage object detectors first extract potential object regions(ie, object proposals) in the first stage. Then in the second stage, the proposal regions are classified into object categories. Earlier approaches extract these RoIs using low-level image features \cite{rcnn-Girshick-2016, fast_rcnn-Girshick-2015}, etc. Later, a learnable component called RPN (Region Proposal Network) is proposed for object proposal extraction, giving birth to end-to-end two-stage detectors~\cite{faster_rcnn-Ren-2015, fpn-Lin-2017, rfcn-Dai-2016}. One-stage object detectors in contrast avoid the RoI extraction stage and classify and regress directly from the anchor boxes. They are  generally fast and applicable to real-time object detection \cite{yolo_9000-Redmon-2017, yolo-Redmon-2016, ssd-Liu-2016, retinanet-Lin-2017}. But generally speaking, two-stage detectors are more accurate and hence used more in aerial detection~\cite{glsan-Deng-2020, dmap-Li-2020}. Recently, one-stage detectors are being explored in aerial images~\cite{querydet-Yang-2022}. These days, anchor-free detectors~\cite{fcos-Tian-2019, centernet-Duan-2019} is getting popular, since they avoid the need for hand-crafted anchor box dimensions and their matching process, which is often specifically tuned depending on the size distribution of the objects. We performed our empirical study mostly on the two-stage detector for a fair comparison with existing approaches, but also report results on the modern anchor-free one-stage detector to validate the generality of our approach.

%\subsection{Detection of Small Objects}
\noindent \textbf{Detection of Small Objects.} 
Small object detection is studied extensively in the literature. Cheng et al. \cite{sod_survey-Cheng-2022} provides a comprehensive survey of small object detection techniques which is broadly classified into: scale aware training \cite{trident_nw-Li-2019, fpn-Lin-2017, snip-Singh-2018, sniper-Singh-2018}, super-resolution methods \cite{pgan-Li-2017}, context modeling \cite{inside_out_net-Bell-2016}, and density guided detection \cite{dmap-Duan-2021, dmap-Li-2020, clusnet-Yang-2019} among others. Due to the crowded distribution of small objects at sparse locations in the high-resolution images, density cropping is a popular strategy in aerial image detection \cite{dmap-Duan-2021, clusnet-Yang-2019, dmap-Li-2020, glsan-Deng-2020}. But this is usually obtained by additional learnable components and multi-stage training. Thus, such methods are more complex to use than uniform cropping. We also use the density cropping strategy but re-purpose the detector itself for extracting the crops, so no additional learnable components are needed. Hence, we believe that our method has the practical simplicity of the uniform cropping  strategy, yet retains the benefit of density guided detection.

%% file: 3-proposal.tex
% !TEX root=main.tex
\section{Proposed Method}
\label{sec:proposal}

This section provides details on how to transform any detector such that it can efficiently process high-resolution images with many small target objects, leveraging density crops. Our proposed cascaded zoom-in (CZ) object detector allows to re-purpose the detector to extract density crops along with base class objects.  Let us consider the \emph{original image}, which is kept at its high resolution, the \emph{down-sampled image}, which is an image containing the same view of the original, but down-scaled to the detector input resolution, and the \emph{cropped images}, which are the selected regions of the image that are up-scaled to the detector resolution. The first component of our pipeline is the density crop labeling algorithm that labels the crowded object regions as "density crops" and augments the training data by adding up-scaled versions of those regions. Then, the density crops are also added to the original image as a new class to be detected. Thus, the augmented training set will include high-resolution versions of the regions containing crowded small objects, allowing the detector to see those small objects in high resolution. Once the model is trained, the inference consists of the detection of the base classes and "density crop" class on the original image, and then a second detection on the predicted density crops after up-scaling. Finally, the detection from the original image and density crops are combined to produce the output detection.

%%%%%%%%
%\subsection{Annotations}
%\subsection{Density Crop Labeling Algorithm}
\subsection{Training with Density Crops}
In order to use a standard detector for our approach, we need to add a new class that we call "density crop" to the training annotations. In this way, our approach is detector agnostic(as we don't change the internals of the detector, we just add one additional class to the list of target classes) and does not require any additional component than the detector itself. The density crop class should label those parts of the image that contains many small objects and include them in a bounding box. This will allow training and inference to focus on those parts by analyzing them in higher resolution. Several different ways could be considered for defining the density crop. The quality constraints we used to define density crops are: (i) they should enclose groups of small target objects,  (ii) they are easy to localize at inference time, and (iii) they are optimal in number to reduce the computational cost.

%Two parameters should be tuned
%\subsection{Density Crop Labeling Algorithm}
Algorithm \ref{alg:crop_discovery} describes the procedure we used for discovering and labeling density crops from the groundtruth(GT) annotations. In summary, we perform an iterative merging of the GT boxes to discover the density crops. In the first step, all GT boxes $\mathcal{B}$ are scaled by expanding the min and max coordinates of the boxes by  $\sigma$ pixels (\verb|scale|($\mathcal{B},\sigma$)). Then we calculate the pairwise Intersection over Union (IoU) between the scaled boxes (\verb|pairwise_IoU|($\mathcal{D}$)) in $O$ as a $|\mathcal{D}| \times|\mathcal{D}|$ matrix. Connections are labeled in $C$ by assigning one to all overlap values above a threshold $\theta$ in the pairwise IoU matrix $O$. Then we select in $C$ the row $m^*$ with the maximum number of connections. %and find all of its connected boxes to form the crop members. 
An enclosing box is computed (\verb|enclosing_box|($C_{m^{*}}$)) by finding the min and max coordinates of all crop members connected to $m^*$. The newly obtained crop box is added to the list of crops and the row $C_{m^*}$ is set to zeros. Subsequently, the crops that are bigger than a maximum threshold $\pi$ are removed from the list $\mathcal{D}$ (\verb|filter_size|($\mathcal{D},\pi$)). This procedure of iterative merging is performed $N$ times. The crop size threshold $\pi$ used here is the ratio of  the area of the crop to that of the image.

%(we used $N=2$ by visually inspecting the quality of the crops; more details are provided in the supplementary material).

% consistent signal as a crop, unlike other methods.

\begin{comment}
%old verison
\begin{algorithm}[t]
    \KwIn{GT boxes in an image: $(\mathcal{B})$}
    \Parameter{$N$: no. of merging steps, $\theta$: overlap threshold}
    \KwOut{Crop boxes}
    1. $\mathcal{B}$'  $\leftarrow$ scale($\mathcal{B}$,{\color{red}S})\;
    2. \For{$i\gets1$ \KwTo $N$ }{
        a) overlaps = pairwise\_IOU($\mathcal{B}$')\\
        b) connections = (overlaps $>$ {\color{red} $\theta$})\\
        c) max\_connected = $\argmax$ rowsum (connections)\\
        d) crop\_members = connections[max\_connected]\\
        e) crop\_box = enclosing\_box(crop\_members)\\
        f) prune crop\_members and update connections\\
        g) $\mathcal{B}' \leftarrow$ filter\-size$(\mathcal{B}',{\color{red}P})$
    }
    \caption{Density Crop Labeling Algorithm.}
\label{alg:cluster_discovery}
\end{algorithm}
\end{comment}

\begin{algorithm}[t]
    \KwIn{$\mathcal{B}$: GT boxes in an image}
    \KwOut{$\mathcal{D}$: Density crops}
    \Parameter{$N$: no. of merging steps, \\${\sigma}$: expansion pixels, \\${\theta}$: overlap threshold, \\${\pi}$: maximum crop size}
    1. $\mathcal{D}  \leftarrow$ {\texttt{scale}}($\mathcal{B},{\sigma}$)\;
    2. \For{$i\gets1$ \KwTo $N$ }{
        %a) $O$ = pairwise\_IOU($\mathcal{B}^*$) \#overlap matrix \\%($|\mathcal{C}| \times |\mathcal{C}|$)\\
        %b) $C$ = $O$ $>$ {\color{red} $\theta$} \ \ \#connection matrix \\%$(K \times K)$ \\
        %a) $O$ = pairwise\_connection($\mathcal{B}^*,\theta$) \\%\#connection matrix \\%
        a) $O$ = {\texttt{pairwise\_IoU}}($\mathcal{D}$)\\
        b) $C = O > \theta$ \\
        c) $\mathcal{D} \leftarrow \emptyset$\\ 
        \While{$|C|>0$}{
        i) $m^* = \argmax_{m} \left( \sum_i C_{m,i} \right)$ \\ %\#max connections \\
        ii) $d$ = {\texttt{enclosing\_box}}($C_{m^*}$) \\%\#new density crop \\        
        %iii) $\mathcal{B}^* \leftarrow$ $\mathcal{B}^*-link(C_m)$\\
        %iv) $C \leftarrow$ $C - link(C_m)$\\
        iii) $\mathcal{D} \leftarrow$ $\mathcal{D}+ d $ \\
        iv) $C_{m^*}$ = 0 \\%\vec{0}$\\
        }
        d) $\mathcal{D} \leftarrow$ {\texttt{filter\_size}}$(\mathcal{D},\pi)$\\
    }
    \caption{Density Crop Labeling Algorithm.}
\label{alg:crop_discovery}
\end{algorithm}

The quality of the crops is important for our method. It is in fact the iterative merging that brings out the best quality crops. Naive scaling and merging to find the maximum enclosing boxes based on pairwise IoU results in either bad crops or too many small crops (with fewer objects in them) depending on the value of the scaling factor. Iterative merging produces good-quality crops enclosing groups of small objects respecting the quality constraints. In section \ref{sec:experiments}, we present the ablation studies validating the effectiveness of our density crop labeling algorithm. We also show that hyperparameters of the algorithm can b easily set. Note that by labeling the density crops apriori,  we are giving the detector a consistent signal of what constitutes a "density crop" throughout the training period, unlike other methods where density crops are also evolving while the training is progressing\cite{clusnet-Yang-2019, dmap-Li-2020, dmap-Duan-2021}. This is consistent with the observations in \cite{detic-zhou-2022} that simple heuristic methods that can give a consistent groundtruth label during training are superior to complex prediction-based groundtruth labels that are continuously evolving.

%\subsection{Training}
With the newly obtained crop labels, we can also augment the training set with additional image crops. The original image and its annotations $\mathcal{B}$ are down-scaled using the maximum training resolution $W \times H$. Note that it is expected the detector will not detect many small objects in the down-scaled image. But the augmented up-scaled version of the density crop $d \in \mathcal{D}$ of a given image will have those small objects that fall inside the crop in a higher resolution. This will reduce the extreme scale variation at training time. The crop labeling can be performed as a pre-processing step. The up-scaled version  augmentation of density crops is simply a data augmentation process. Thus our method is not introducing any change in the standard training pipeline of a detector, except the new class "density crop" is added. In this regard, it is practically easy to use like uniform cropping.

%%%%
\subsection{Multi-stage Inference}
As the detector is trained to recognize density crops, at inference, we can get the density crop from its prediction itself. Figure~\ref{fig:base_daigram} bottom explains our inference process in detail. It consists of two stages. In stage one, it predicts the base class objects and density crops on the input image. Then we select the high-quality density crops based on their confidence score. In stage two, the upscaled density crops are passed through the same detector again, producing small object detection on the density crops. Finally, we re-project the detections on the crops to the original image and concatenate them with the detections on the original image. 
Let $c \in \mathcal{C}$ be an up-scaled crop image of size $(I^W_c,I^H_c)$ defined by its bounding box coordinates $(c_{x1}, c_{y1}, c_{x2}, c_{y2})$ in the original image. Given the scaling factors $(S^W_c,S^H_c)=(\frac{c_{x2}-c_{x1}}{I^W_c},\frac{c_{y2}-c_{y1}}{I^H_c})$, the re-projection box $p_i$ scales down and shifts the detection boxes  $(x_{1,i},y_{1,i},x_{2,i},y_{2,i}) \in \mathcal{B}^c$ in the crop $c$ as:
\begin{align}
    %& p_i = \frac{y^c_2-y^c_1}{I_H}b_i^c + (x^c_1, y^c_1, x^c_1, y^c_1) 
    p_i = & ({S^W}x_{1,i},{S^H}y_{1,i},{S^W}x_{2,i},{S^H}y_{2,i}) \notag \\
    & + (c_{x1}, c_{y1}, c_{x1}, c_{y1}) 
    %& with \hspace{3pt} b_i^c \in \mathcal{B}^c.
    \label{equ:reproj}
\end{align}
\begin{comment}
Let $c \in \mathcal{C}$ be an up-scaled crop with height $c^U_H$, whose height  and position in the input image $I$ are $c^I_H$ and $[x^c_1, y^c_1, x^c_2, y^c_2]$ respectively. The re-projection operation simply scales down and shifts the detection boxes $\mathcal{B}^c$ in $c$ as shown in equation \ref{equ:reproj}:
\begin{equation}
    \mathcal{B}^c_{proj} = \frac{c^I_H}{c^U_H} \mathcal{B}^c \oplus [x^c_1, y^c_1, x^c_1, y^c_1].
    \label{equ:reproj}
\end{equation}
where $\oplus$ is an element-wise addition operation. 
\end{comment}
The Non-Maximal Suppression(NMS) is then applied to remove duplicate detections.

While other methods need complex post-processing to filter the noisy crops\cite{clusnet-Yang-2019}, we can simply use the confidence score of the density crops to do the same. Stage one of the inference is the standard inference procedure in any detector. The filtering of the noisy crops can be easily performed with the confidence scores given by the detector. The second stage of the inference is performed with the same detector, but a different input (the up-scaled density crops). So, we are simply repeating the standard inference procedure of a detector one more time. All of these operations can be easily wrapped on top of the inference procedure of any detector, thus keeping the simplicity of the uniform cropping approach at inference too.

%% file: 4-experiments.tex
% !TEX root=main.tex

\section{Experiments}
\label{sec:experiments}

%%
%\subsection{Experimental Setup}
\noindent \textbf{Datasets and evaluation measures.} For evaluation of methods, we employed two popular challenging benchmark datasets for Aerial Image Detection, namely the VisDrone \cite{zhu-VisDrone-2018} and DOTA \cite{xia-DOTA-2018} datasets. The measure used for assessing and comparing the performance of methods is COCO style average precision (AP) \cite{mscoco-Lin-2014}. The AP of small, medium and large objects are also reported, particularly to understand the performance of our method for small object detection. Finally, the number of frames per second (FPS) is reported as a measure of time complexity.

\noindent \textbf{VisDrone.} This dataset contains  8,599 drone-captured images (6,471
for training, 548 for validation, and 1,580 for testing) with a resolution of about 2000 $\times$1500 pixels. The objects are from ten categories with 540k instances annotated in the training set, mostly containing different categories of vehicles and pedestrians observed from drones. It has an extreme class imbalance and scale imbalance making it an ideal benchmark for studying small object detection problems. As the evaluation server is closed now, following the existing works, we used the validation set for evaluating the performance.

\noindent \textbf{DOTA.} This dataset is comprised of satellite images. The images in this dataset have a resolution ranging from 800$\times$800 to 4000$\times$4000. Around 280k annotated instances are present in the dataset. The objects are from fifteen different categories, with movable objects such as planes, ships, large vehicles, small vehicles, and helicopters. The remaining ten categories are roundabouts, harbors, swimming pools, etc. Many density crop based detection papers reports results only on movable objects\cite{clusnet-Yang-2019} with the assumption that immovable objects usually won't appear crowded. But they are also small objects, so we kept all classes to assess the improvement in small object detection. The training and validation data contain 1411 images and 458 images, respectively.

\noindent \textbf{Implementation details.} The Detectron2 toolkit \cite{detectron2-wu-2019} was used to implement our CZ detector. The backbone detector used in our study is primarily Faster RCNN \cite{faster_rcnn-Ren-2015}, but we also show results on the modern anchor-free one-stage detector FCOS \cite{fcos-Tian-2019}. This validates our claim that our approach gives a consistent improvement in performance regardless of the detector used. We used Feature Pyramid Network (FPN) \cite{fpn-Lin-2017} backbone with ResNet50 \cite{resnet-He-2016}  pre-trained on ImageNet \cite{imagenet-Russakovsky-2015} dataset for our experimental validation. For data augmentation, we resized the shorter edge to one randomly picked from (800, 900, 1000, 1100, 1200), and applied horizontal flip with a 50\% probability. The model was trained on both datasets for 70k iterations. The initial learning rate is set to 0.01 and decayed by 10 at 30k and 50k iterations. For training, we used one NVIDIA A100 GPU with 40 GB of memory.

%%%%%%%%
\subsection{Comparison with Baselines}

\begin{figure*}
\centering
\begin{tabular}{c @{\hspace{-0.3cm}} c  @{\hspace{-0.3cm}} c}
\includegraphics[width=0.32\linewidth, height=3cm]{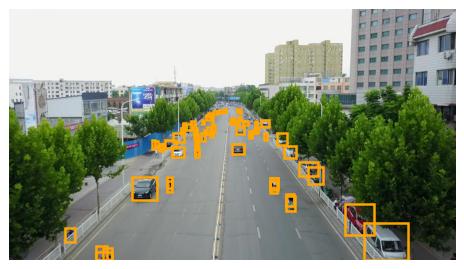} &
\includegraphics[width=0.32\linewidth, height=3cm]{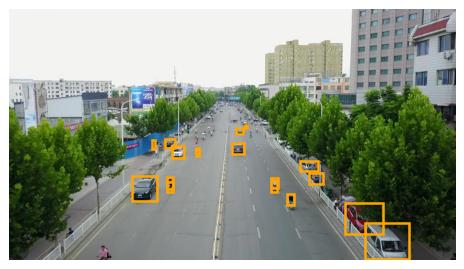} &
\includegraphics[width=0.32\linewidth, height=3cm]{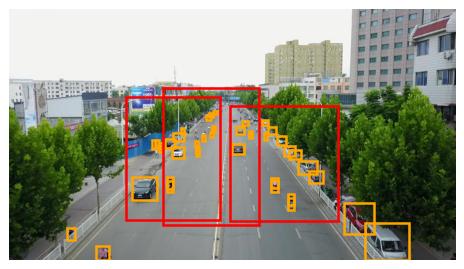} \\
\includegraphics[width=0.32\linewidth, height=3cm]{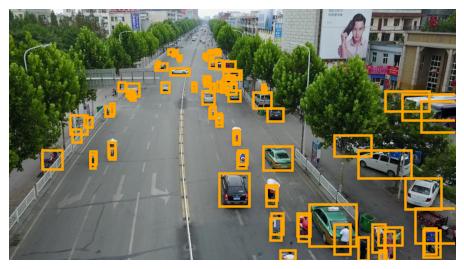} &
\includegraphics[width=0.32\linewidth, height=3cm]{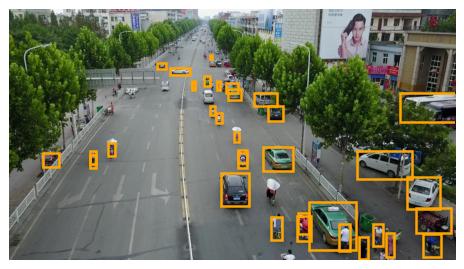} &
\includegraphics[width=0.32\linewidth, height=3cm]{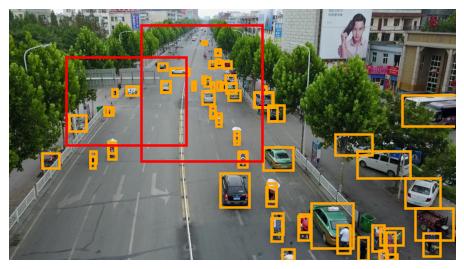} \\
\includegraphics[width=0.32\linewidth, height=3cm]{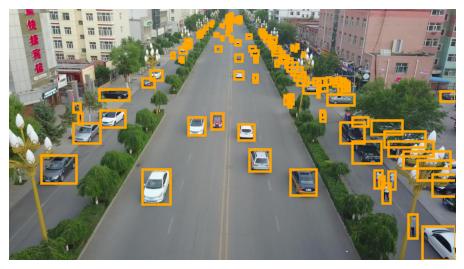} &
\includegraphics[width=0.32\linewidth, height=3cm]{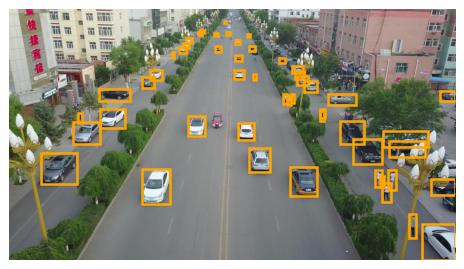} &
\includegraphics[width=0.32\linewidth, height=3cm]{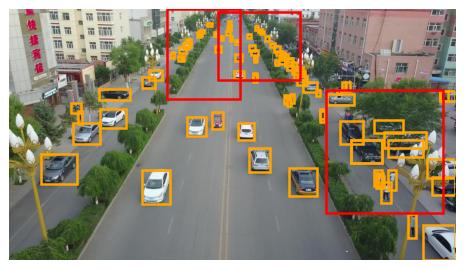} \\
 (a) image \& GT & (b) baseline detection & (c) detection with crops \\
 \end{tabular}
 \caption{Visualization of density crop-based detection. (a) the original image and its GT. (b) detection with the baseline detector. (c) detection with density crops; the density crops are shown in red color. Our method detects more objects, especially inside the crop regions.}
 \label{fig:clusters}
\end{figure*}

%\subsubsection{Uniform Crops vs Density-based Crops}
Table \ref{table:comparison_with_uniform_crops} presents a comparison between uniform cropping and density cropping on the VisDrone dataset, with and without the last feature map of the feature pyramid (P2), which has a strong impact in memory and computation \cite{querydet-Yang-2022}. For the uniform cropping, we crop the original image into 4 equal-sized crops by splitting at half height and width. In order to have a fair comparison, we use our method with a confidence threshold of 0.7 to obtain an average of  1-3 crops per image. The observations in the table suggest that uniform cropping improves performance compared to vanilla training on the whole image, but it is still inferior to our density-based cropping. When high-resolution feature maps P2 are not used, density cropping gains more than 3.5 points in AP and the AP of small objects is improved by 3.4 points. It is worth noting that compared to uniform cropping, our approach introduces additional parameters to recognize one extra class and no changes in learning and inference dynamics. So this can be easily used as a plug-and-play replacement for the uniform crop-based training, popular among the community. In terms of frame rate, our approach is slightly slower than uniform crops. However, we observe that our method without the expensive P2 features performs better than uniform crops with P2, while also being faster. In Figure \ref{fig:clusters}, a visual comparison of the highly confident detections between the baseline model and our density crop-based model is shown. When the density crops are used, we can observe an increase in the number of detections. It can be observed that more objects are getting discovered in the crop regions when the detection results from the second inference are augmented. This explains the impact of our zoom-in detector for small object detection in high-resolution images.

\setlength{\textfloatsep}{0.3cm}
\begin{table}
    \centering
    \resizebox{.48\textwidth}{!}{% <------ Don't forget this %
      \begin{tabular}{l||rrr|rrr|r}
    \textbf{Settings} & \textbf{AP} & \textbf{$\textrm{AP}_{50}$} & \textbf{$\textrm{AP}_{75}$} & \textbf{$\textrm{AP}_{s}$}  & \textbf{$\textrm{AP}_{m}$} & \textbf{$\textrm{AP}_{l}$} & \textbf{FPS}\\
    \hline
    \emph{Without P2} & & & & & & & \\
    Baseline  & 29.48 & 51.68 & 29.55 & 22.33 & 38.66 & 39.30 & 26.31\\
    Uniform crops & 30.68 & 54.44 & 30.54 & 22.91 & 40.62 & 41.03 & 12.30\\
    CZ Det. (ours) & 33.02 & 57.87 & 33.09 & 25.74 & 42.93 & 41.44 & 11.64\\
    \hline    
    \emph{With P2} & & & & & & & \\
    Baseline & 30.81 & 55.06 & 30.68 & 23.97 & 39.19 & 41.17 & 18.25\\    
    Uniform crops & 31.73 & 56.31 & 31.57 & 25.13 & 40.41 & 41.06 & 9.85\\
    CZ Det. (ours) & 33.22 & 58.30 & 33.16 & 26.06 & 42.58 & 43.36 & 8.44\\
    
    \end{tabular}% <------ Don't forget this %
    }
    \caption{Comparison of detection performance between a baseline detector, uniform crops and density crops on VisDrone dataset (1.5K pixel resolution). %Uniform cropping creates further scale imbalance and decreases performance.
    }
    \label{table:comparison_with_uniform_crops}
\end{table}

To further verify the observations, we repeated the same type of study in the satellite images of the DOTA dataset. In this dataset images are at higher resolution(4k pixels), thus due to memory constraints, the baselines are already performing uniform cropping. Table \ref{table:dota_results} shows the results of a uniform cropping baseline and our CZ detector for two different configurations. Similar to VisDrone, significant improvement is seen in the case of not using high-resolution features P2, with a gain of 2.9 points. APs of small and medium objects are improved by 3.0 and 4.0 points respectively from the baseline without using high-resolution features. In terms of computation, we can see that, as expected, our approach has a slightly slower frame rate than the baseline. However, this is compensated by the higher detection accuracy. We see for instance that the best baseline with P2 features has an AP of $33.44\%$ with an FPS of $0.49$, while our CZ detector without P2 features has a higher AP ($34.14\%$) while being also faster (0.62 FPS).
%Then the second (2k $\times$ 2k without P2) up-scales the image crops at higher resolution while still not detecting on P2. 
%We can also conclude that with our approach the gap between with and without high resolution features is minimized, signifying its boost in small object detection performance. Thus we can avoid using expensive high-resolution features without compromising the detection accuracy.
\begin{table}
    \centering
    \resizebox{.48\textwidth}{!}{% <------ Don't forget this %
      \begin{tabular}{l||rrr|rrr|r}
    \textbf{Settings} & \textbf{AP} & \textbf{$\textrm{AP}_{50}$} & \textbf{$\textrm{AP}_{75}$} & \textbf{$\textrm{AP}_{s}$}  & \textbf{$\textrm{AP}_{m}$} & \textbf{$\textrm{AP}_{l}$} & 
    \textbf{FPS} \\  
    \hline
    \emph{Without P2} & & & & & &\\
    Baseline  & 31.29 & 51.57 & 33.10 & 12.69 & 34.04 & 42.83 & 0.93 \\
    CZ Det. (Ours) & 34.14 & 56.69 & 35.69 & 15.66 & 38.16 & 44.20 & 0.62 \\
    \hline
    %\emph{2k$\times$2k without P2} & & & & & &\\
    %Baseline & 32.73 & 52.94 & 34.96 & 16.71 & 34.92 & 43.74 & 0.54 \\ 
    %CZ Det. (Ours) & 34.52 & 56.57 & 36.41 & 15.85 & 37.26 & 45.32 & 0.35 \\
    %\hline
    \emph{With P2} &&&&&& \\
    Baseline & 33.44 & 54.03 & 35.56 & 16.86 & 36.76 & 43.65 & 0.49 \\
    CZ Det. (Ours) & 34.62 & 56.86 & 36.17 & 18.17 & 37.84 & 43.83 & 0.30 \\
    \end{tabular}% <------ Don't forget this %
    }
    \caption{Performance comparison of our method against baselines on DOTA dataset (4K pixel resolution).}
    \label{table:dota_results}
\end{table}

%%%%%
\subsection{Ablation Studies}

%In this section, we present the observations from several ablation studies conducted to validate the effectiveness of our approach. We used the VisDrone dataset for all studies in this section. Additional studies about the detector backbone, the analysis of the detection errors, and the effect of the maximum crop size $\pi$ are presented in the supplementary material.

The effectiveness of our proposed CZ detector is characterized by ablation experiments on the VisDrone dataset. Additional studies showing the impact of hyperparameters in the crop labeling algorithm, different backbone networks for the detector, and an analysis of the detection errors are presented in the supplementary material.

\noindent \textbf{a) Density Crops effect at Training and Inference.}  
%\paragraph{Density Crops effect at Training and Inference.}
We used density crops at the training and test time to achieve optimal performance. In particular, while training, the rescaled density crops are augmented with the training images; while testing we do the two-stage inference where stage one perform inference on the whole image and stage two performs inference on the density crops. In this section, we study the importance of this configuration. Table \ref{table:with_and_without_crops} shows the results. When the density crops are not augmented with the training set but only used in the two-stage inference, the improvement is marginal over the baseline(most importantly, $\textrm{AP}_{s}$ has no change). This is because the scale imbalance in the input image is not mitigated as the detector is not seeing the small objects at a bigger scale. When density crops are added to the training set, the detection accuracy improves significantly. However, the inference is still happening on the whole image so the detection accuracy of small objects is affected. When inference is performed on the density crops and fused with the detection on the whole image, we get the best results.
\setlength{\textfloatsep}{0.3cm}
\begin{table}
    \centering
    \resizebox{.48\textwidth}{!}{% <------ Don't forget this %
      \begin{tabular}{rr||rrr|rrr}
    \textbf{Train} & \textbf{Test} & \textbf{AP} & \textbf{$\textrm{AP}_{50}$} & \textbf{$\textrm{AP}_{75}$} & \textbf{$\textrm{AP}_{s}$}  & \textbf{$\textrm{AP}_{m}$} & \textbf{$\textrm{AP}_{l}$} \\  
    \hline
     &  & 29.48 & 51.68 & 29.55 & 22.33 & 38.66 & 39.30 \\
      & $\checkmark$ & 29.93 & 53.29 & 29.52 & 22.33 & 39.35 & 39.46 \\
    $\checkmark$ &  & 32.64 & 57.36 & 32.78 & 24.81 & 43.04 & 41.07 \\
    $\checkmark$ & $\checkmark$ & 33.02 & 57.87 & 33.09 & 25.74 & 42.93 & 41.44 \\
    \end{tabular}% <------ Don't forget this %
    }
    \caption{Detection results with and without density crops at train time and test time.}
    \label{table:with_and_without_crops}
\end{table}

\begin{figure}[h!]
  \centering
  \includegraphics[scale=0.39]{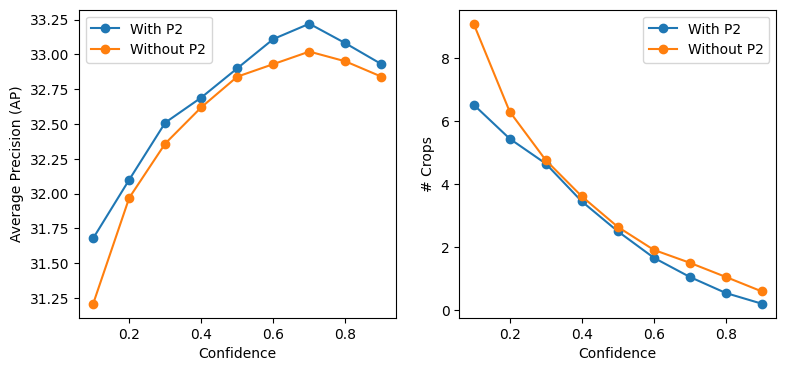} 
  \caption{Change in detection precision and the number of crops according to crop confidence. The crop confidence is varied from 0.1 to 0.9. The crop confidence for best detection accuracy is 0.7.}
  \label{fig:conf_impact}
\end{figure}

\noindent \textbf{b) Impact of the Quality of Crops.}  
%\paragraph{Impact of the Quality of Crops.}
Figure \ref{fig:conf_impact} illustrates how the confidence of crops impacts the detection accuracy and the number of density crops extracted. The impact is studied for two settings, with and without the high-resolution features P2. This is to verify how the density crops aid small object detection with and without utilizing expensive high-resolution feature maps. The crop confidence, which is used as the proxy for crop quality,  is varied from 0.1 to 0.9. In general, with lower confidence values, we are observing more crops but many of them are noisy and redundant even after Non-Maximal Suppression. So when the quality of the crops is low, the detection accuracy decreases (Figure \ref{fig:conf_impact} left). When the quality is increased, the accuracy increases until 0.7, and then it is gradually coming down as we use very few crops in that case. The trend is the same with and without P2.

From Tables \ref{table:comparison_with_uniform_crops} and \ref{table:dota_results}, we observed that density crops obtained better gain in detection accuracy over the baseline without high-resolution features. Though this is expected, we decided to understand how exactly this is happening. We analyzed the number of density crops retained after filtering out the low-quality crops at multiple confidence levels ranging from 0.1 to 0.9. Figure \ref{fig:conf_impact} right shows the results with and without high-resolution features P2. It can be observed that for "without P2", we are getting more density crops at all confidence levels. With higher crop confidence levels, we get more high-quality crops for "without P2" case, hence we observe a better gain in detection accuracy over the baseline. We used a confidence of 0.7 in all our experiments to have the best trade-off between detection precision and speed. While other methods use post-processing on the crop detections \cite{clusnet-Yang-2019} or density maps \cite{dmap-Li-2020} to filter the noisy crops during inference, we can filter them out based on their confidence score simplifying the inference procedure.

\noindent \textbf{c) Why Iterative Merging for Crop Discovery?}  
%\paragraph{Why Iterative Merging for Crop Discovery?}
Simply scaling and doing a one-step merging operation to create density crops results in sub-optimal crops. We empirically verify this with multiple scaling strategies and argue that the iterative merging strategy is superior to them. Authors of \cite{clusnet-Yang-2019} also used iterative crop merging on the output of their crop detection module to reduce the redundant crops. This has to be performed at training and test time to refine the initial crop detections. To label the crops for training, they used a single-step aggregation. Our iterative merging for labeling crops can be performed as a pre-processing step before training. We avoided redundant crops at inference, by filtering them out based on the confidence score.

Table \ref{table:scale_by_factor} top provides the comparative results of single-step merging with our iterative merging strategy where GT boxes are scaled by a scaling factor. Using a low scaling factor creates too many crops, containing fewer objects. More specifically, it is producing multiple small crops containing fewer objects in crowded regions in the image. When the scaling factor is increased, the number of crops decreases and performance increases up to a point but declines later as the crops become too big and the object density of the image is less respected. This is because large scaling factors significantly blow up the big GT boxes and it alters the density of the crops. The detection performance obtained is also far below our iterative merging. Table \ref{table:scale_by_factor} bottom shows the same comparison when GT boxes are scaled by constant pixel values. As this avoids the blowing of large bounding boxes due to the constant scaling, the detection performance is better than the former one. Iterative merging produces the optimal number of crops with the best performance. The scaling used in the iterative merging is small and only performed at the first stage of merging. We used 20 pixels as the scaling magnitude. Large values are not possible here since the \texttt{filter\_size} operation while restricting the crop size will reduce the number of crops. Thus it is easy to set. In the supplementary material, we provide more experiments studying the ease of tuning the hyperparameters $N, \pi, \theta$ of the crop labeling algorithm, and the visualization of the crops obtained by different merging strategies.

\begin{table}
    \centering
    \resizebox{.48\textwidth}{!}{% <------ Don't forget this %
      \begin{tabular}{r||r|rrr|rrr}
    \textbf{Scaling} & \textbf{\# crops} &  \textbf{AP} & \textbf{$\textrm{AP}_{50}$} & \textbf{$\textrm{AP}_{75}$} & \textbf{$\textrm{AP}_{s}$}  & \textbf{$\textrm{AP}_{m}$} & \textbf{$\textrm{AP}_{l}$} \\  
    \hline  
    Baseline & 0 & 29.48 & 51.68 & 29.55 & 22.33 & 38.66 & 39.30 \\
    \hline
    factor = 2.0 & 74417 & 24.39 & 44.38 & 23.73 & 15.96 & 34.96 & 47.24 \\
    3.0 & 67906 & 30.64 & 53.71 & 30.68 & 23.23 & 40.64 & 39.35 \\
    6.0 & 43300 & 31.30 & 55.33 & 31.42 & 23.79 & 41.06 & 38.40 \\
    8.0 & 34663 & 30.95 & 55.18 & 30.31 & 23.38 & 40.93 & 39.24 \\
    \hline
    pixels = 30 & 62677 & 31.26 & 54.55 & 31.50 & 23.83 & 40.78 & 50.07 \\
    60 & 46753 & 31.98 & 55.84 & 32.07 & 25.12 & 41.11 & 45.52 \\
    90 & 35442 & 31.47 & 55.62 & 30.96 & 24.03 & 41.27 & 44.08 \\
    120 & 25146 & 31.07 & 54.74 & 30.95 & 23.18 & 41.43 & 42.39 \\
    \hline
    Ours & 14018 & 33.02 & 57.87 & 33.09 & 25.74 & 42.93 & 41.44 \\
    \end{tabular}% <------ Don't forget this %
    }
    \caption{Comparison of iterative merging strategy with single-step merging where GT boxes are scaled according to scaling factors, and scaled uniformly by pixel values.}
    \label{table:scale_by_factor}
\end{table}

\subsection{Results with Other Detectors}
To validate the effectiveness of our approach with other detection architectures, we conducted experiments on the modern anchor-free one-stage detector FCOS \cite{fcos-Tian-2019}. Table \ref{table:fcos_result} shows the performance comparison of the vanilla FCOS detector with our density crop based FCOS detector. Similar to the results in table \ref{table:comparison_with_uniform_crops}, \textrm{AP} is improved by a significant margin, and $\textrm{AP}_{s}$ has gained almost 5 points. 

\begin{table}
    \centering
    \resizebox{.48\textwidth}{!}{% <------ Don't forget this %
      \begin{tabular}{r||rrr|rrr|r}
    \textbf{Settings} & \textbf{AP} & \textbf{$\textrm{AP}_{50}$} & \textbf{$\textrm{AP}_{75}$} & \textbf{$\textrm{AP}_{s}$}  & \textbf{$\textrm{AP}_{m}$} & \textbf{$\textrm{AP}_{l}$} & \textbf{FPS} \\  
    \hline
    Base FCOS & 29.51 & 50.40 & 29.92 & 21.25 & 40.51 & 37.29 & 26.01\\
    CZ FCOS Det. & 33.67 & 56.20 & 34.15 & 26.16 & 43.98 & 46.87 & 12.69\\
    \end{tabular}% <------ Don't forget this %
    }
    \caption{Results with anchor free detector FCOS on the Visdrone dataset. All results are without using P2.}
    \label{table:fcos_result}
\end{table}

%%%%%%%%%
\subsection{Comparison with State-of-the-Art Methods}
Table \ref{table:visdrone_sota_comparison} compares our approach with the existing methods on the VisDrone dataset. Similarly to us, some methods perform density cropping \cite{clusnet-Yang-2019,dmap-Li-2020,dmap-Duan-2021,glsan-Deng-2020}, while QueryNet\cite{querydet-Yang-2022} and CascadeNet \cite{cascadenet-Zhang-2019} use other approaches to improve the detection performance on aerial images. We obtained the best detection AP among the state-of-the-art methods. Only for large objects, DensityMap performs better than our approach. This is probably because our method gets biased to detect small objects, thanks to the additional crops on training. In fact, for small object detection, we obtained the best $\textrm{AP}_{s}$, significantly outperforming all existing approaches. $\textrm{AP}_{m}$ also shows a good improvement of more than 2 points.

\begin{table}
    \centering
    \resizebox{.48\textwidth}{!}{% <------ Don't forget this %
      \begin{tabular}{l||rrr|rrr}
    \textbf{Method} & \textbf{AP} & \textbf{$\textrm{AP}_{50}$} & \textbf{$\textrm{AP}_{75}$} & \textbf{$\textrm{AP}_{s}$}  & \textbf{$\textrm{AP}_{m}$} & \textbf{$\textrm{AP}_{l}$} \\ 
    \hline   
    ClusterNet \cite{clusnet-Yang-2019} & 26.72 & 50.63 & 24.70 & 17.61 & 38.92 & 51.40 \\
    DensityMap \cite{dmap-Li-2020} & 28.21 & 47.62 & 28.90 & 19.90 & 39.61 & \textbf{55.81} \\
    CDMNet \cite{dmap-Duan-2021} & 29.20 & 49.50 & 29.80 &  20.80 & 40.70 & 41.60 \\
    GLSAN \cite{glsan-Deng-2020} & 30.70 & 55.40 & 30.00 & - & - & - \\ 
    QueryDet \cite{querydet-Yang-2022} & 28.32 & 48.14 & 28.75 & - & - & - \\
    CascadeNet \cite{cascadenet-Zhang-2019} & 28.80 &  47.10 & 29.30 & - & - & - \\
    CascNet+MF \cite{cascadenet-Zhang-2019} & 30.12 &  58.02 & 27.53 & - & - & - \\    
    \hline 
    %Ours(without P2) & 33.02 & 57.87 & 33.09 & 25.74 & \textbf{42.93} & 41.44 \\
    CZ Det. (Ours) & \textbf{33.22} & \textbf{58.30} & \textbf{33.16} & \textbf{26.06} & \textbf{42.58} & 43.36 \\
    \end{tabular}% <------ Don't forget this %
    }
    \caption{Performance of our proposed method compared against state-of-art approaches with Faster RCNN detector on the VisDrone validation set. "MF" stands for model fusion. The best results in each column are highlighted in \textbf{bold}.}
    \label{table:visdrone_sota_comparison}
\end{table}

%% file: 5-conclusion.tex
\section{Conclusion}
\label{sec:conclusion}
We proposed an efficient method for utilizing density crops for aerial image detection. Our method is as simple to use as the uniform cropping approach widely used by practitioners. The training step simply adds an additional class called "density crop" to the detector whose labels are obtained from a crop labeling algorithm. The inference is performed in two steps, one on the original image and then on the up-scaled version of the high-quality crops detected from it. For both modifications, we re-purpose the original detector, thus alleviating the need for additional components, unlike existing approaches. Empirical results verify the superiority of our approach in terms of detection accuracy and ease of use. At present, the density crops are up-scaled to a fixed resolution. In the future, we plan to up-scale adaptively, respecting the density of the crops. 

%% file: 6-supp-materials.tex
% !TEX root=main.tex
\section{Appendix}
This appendix provides additional ablation studies and analyses using the VisDrone dataset.

\subsection{Impact of hyperparameters in the crop labeling algorithm}
In this section, we analyzed the impact of different hyperparameters in the crop labeling algorithm. Subsequently, we observed that they can be easily tuned.
\subsubsection{Impact of $N$ in the crop-labeling algorithm}
\begin{figure*}
\centering
\begin{tabular}{c @{\hspace{-0.2cm}} c  @{\hspace{-0.2cm}} c @{\hspace{-0.2cm}} c}
\includegraphics[width=0.24\linewidth]{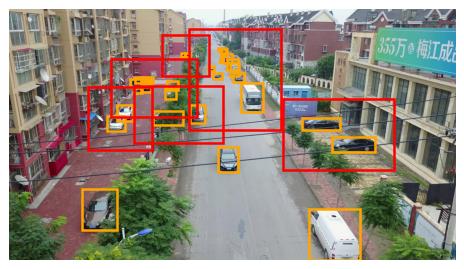} &
\includegraphics[width=0.24\linewidth]{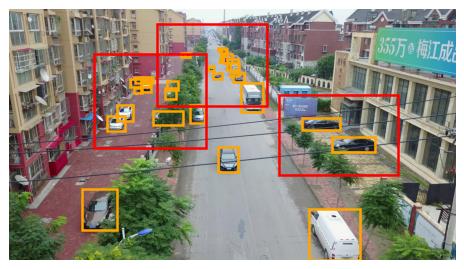} &
\includegraphics[width=0.24\linewidth]{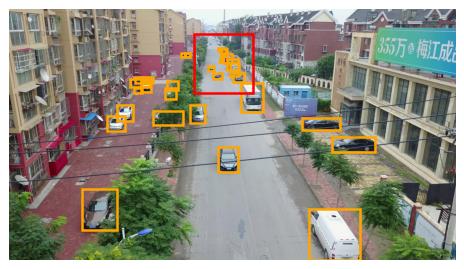} &
\includegraphics[width=0.24\linewidth]{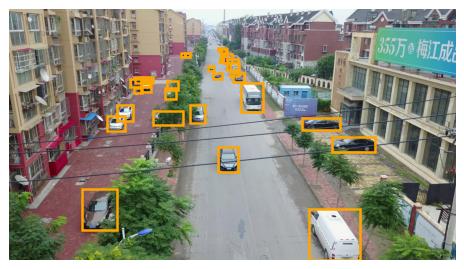} \\
\includegraphics[width=0.24\linewidth]{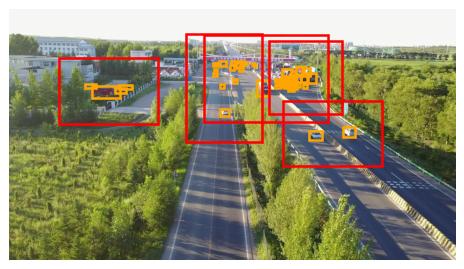} &
\includegraphics[width=0.24\linewidth]{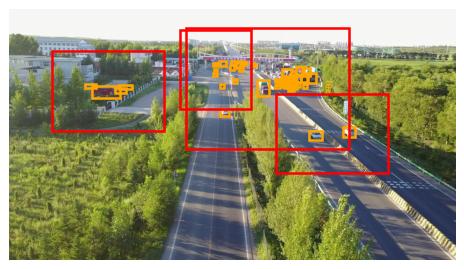} &
\includegraphics[width=0.24\linewidth]{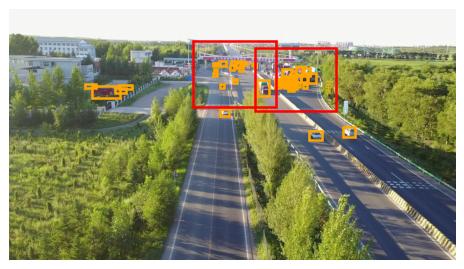} &
\includegraphics[width=0.24\linewidth]{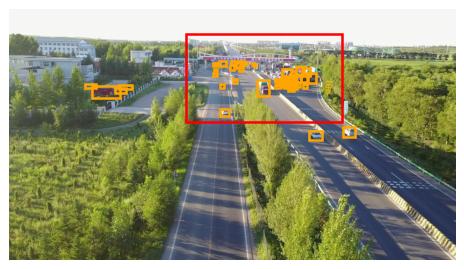} \\
\includegraphics[width=0.24\linewidth]{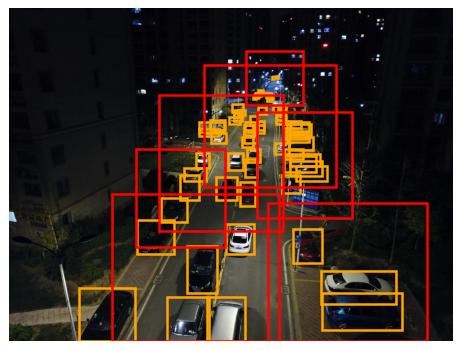} &
\includegraphics[width=0.24\linewidth]{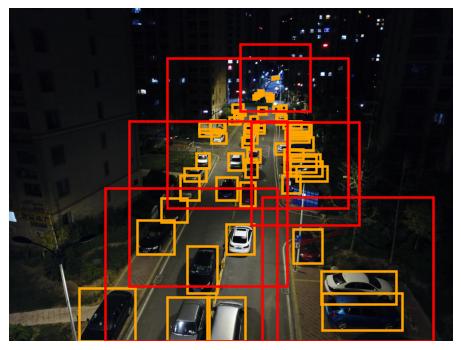} &
\includegraphics[width=0.24\linewidth]{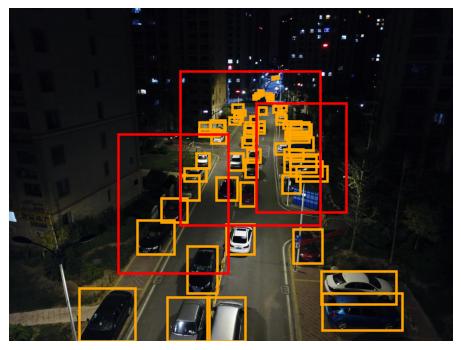} &
\includegraphics[width=0.24\linewidth]{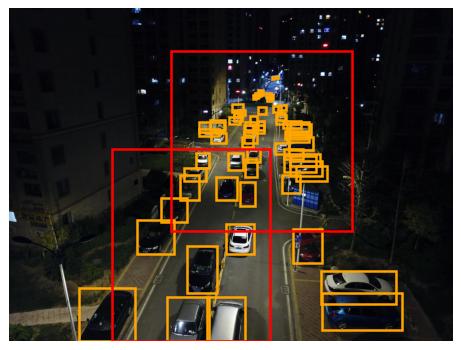} \\
(a) 1-stage pixel & (b) 1-stage factor & (c) 2-stage & (d) 3-stage \\
 \end{tabular}
 \caption{Comparison of different crop-labeling algorithms. (a) 1-stage with scaling by pixel (b) 1-stage with scaling by a scaling factor. (c) our 2-stage iterative merging. (d)3-stage iterative merging. 1-stage merging is producing too many crops even grouping larger objects in the image; some crops contain very few objects. 2-stage merging produces the optimal number of crops by mostly enclosing small objects in the image. 3-stage merging is producing very few crops, often spanning to large areas in the background regions. Sometimes the crops are even disappearing due to the size constraint limiting the crop size(eg: col 4 in the first row).}
 \label{fig:crop-labeling}
\end{figure*}

When we performed iterative merging, we set the number of merging steps $N$=2. In this section, we empirically verify the impact of other possible $N$ values. Table \ref{table:n_merging_impact} shows the comparison. When $N$=1, we have too many small noisy crops. The optimal quality crops are obtained when $N$ = 2. When $N$ = 3, the number of crops reduces, and crops tend to grow significantly to cover many background regions. Thus we see a reduction in performance. A visualization of the crops shows further evidence regarding the quality of crops when iterative merging is used in the crop-labeling. Figure \ref{fig:crop-labeling} shows the comparison when crop-labeling is performed with different iteration values $N$. When $N$ = 1, regardless of whether we scale the boxes by pixel or a scaling factor, we get too many small crops often containing a few objects, violating our quality constraints. It is also producing crops around large objects in the image. $N$ = 2 gives the best quality crops, optimal in number according to the object density, and encloses mostly the small objects. $N$ = 3 enlarges the crops so much covering background regions, also very large crops are getting filtered out(fig \ref{fig:crop-labeling} first row, third column). This establishes that our crop labeling algorithm is respecting the specified quality constraints.

\begin{table}
    \centering
    \resizebox{.48\textwidth}{!}{% <------ Don't forget this %
      \begin{tabular}{r||r|rrr|rrr}
    \textbf{$N$} & \textbf{\# crops} &  \textbf{AP} & \textbf{$\textrm{AP}_{50}$} & \textbf{$\textrm{AP}_{75}$} & \textbf{$\textrm{AP}_{s}$}  & \textbf{$\textrm{AP}_{m}$} & \textbf{$\textrm{AP}_{l}$} \\  
    \hline
    1 & 62677 & 31.26 & 54.55 & 31.50 & 23.83 & 40.78 & 50.07 \\
    2 & 14018 & 33.02 & 57.87 & 33.09 & 25.74 & 42.93 & 41.44 \\
    3 & 2227 & 31.14 & 55.22 & 30.88 & 23.99 & 40.26 & 40.43 \\
    \end{tabular}% <------ Don't forget this %
    }
    \caption{Impact on performance of the number of iterative merging steps $N$ used the density crop labeling algorithm.}
    \label{table:n_merging_impact}
\end{table}

\subsubsection{Impact of the $\theta$ on the Crop Labeling Algorithm.}  
%\paragraph{Impact of the $\theta$ on the Crop Labeling Algorithm.}
In Table \ref{table:theta_impact}, we studied the impact of the $\theta$ parameter in the crop labeling algorithm. From the results, it can be observed that a small value of the $\theta$ is preferred. The bounding boxes of the small objects exhibit low IoU values, hence maximum connections are observed for low values of $\theta$. A large value for the overlap parameter $\theta$ affects the performance significantly. This is expected, as we increase the threshold, the connections in the crop labeling algorithm reduce, and hence the number of density crops discovered also reduces. This will subsequently move towards the baseline case where no density crops are used both at the train and test time.
\begin{table}
    \centering
    \resizebox{.48\textwidth}{!}{% <------ Don't forget this %
      \begin{tabular}{r||r|rrr|rrr}
    \textbf{$\theta$ Value} & \textbf{\# crops} &  \textbf{AP} & \textbf{$\textrm{AP}_{50}$} & \textbf{$\textrm{AP}_{75}$} & \textbf{$\textrm{AP}_{s}$}  & \textbf{$\textrm{AP}_{m}$} & \textbf{$\textrm{AP}_{l}$} \\  
    \hline
    0.1 & 14018 & 33.02 & 57.87 & 33.09 & 25.74 & 42.93 & 41.44 \\
    0.2 & 12652 & 32.80 & 57.62 & 32.82 & 25.50 & 42.43 & 43.56 \\
    0.3 & 10222 & 32.66 & 57.25 & 32.70 & 25.57 & 42.33 & 44.10 \\
    0.4 & 7551 & 32.08 & 56.69 & 31.84 & 24.26 & 42.47 & 43.99 \\    
    0.5 & 4836 & 31.48 & 55.87 & 31.13 & 24.23 & 41.10 & 38.19 \\    
    \end{tabular}% <------ Don't forget this %
    }
    \caption{Impact on performance of the overlap threshold $\theta$ in the density crop labeling algorithm.}
    \label{table:theta_impact}
\end{table}

%%%%
\subsubsection{Impact of $\pi$ in the crop-labeling algorithm}
We used a maximum size limit for the crops to ignore oversized crops. The size parameter $\pi$ in the crop-labeling algorithm represents the ratio of the area of the crop to that of the image and allows us to ignore those crops whose area ratio is above a threshold. This parameter serves only to filter out unusually big crops spanning a big portion of the image and requires minimal tuning. Table \ref{table:max_clus_size_impact} shows the results in the case of the VisDrone dataset. As we can see, there are few crops growing significantly when using a 2-stage iterative merging. So setting $\pi$ = 0.3 provides the best quality crops on average. 
\begin{table}
    \centering
    \resizebox{.48\textwidth}{!}{% <------ Don't forget this %
      \begin{tabular}{r||r|rrr|rrr}
    \textbf{$\pi$} & \textbf{\# crops} &  \textbf{AP} & \textbf{$\textrm{AP}_{50}$} & \textbf{$\textrm{AP}_{75}$} & \textbf{$\textrm{AP}_{s}$}  & \textbf{$\textrm{AP}_{m}$} & \textbf{$\textrm{AP}_{l}$} \\  
    \hline
    0.1 & 11574 & 32.23 & 56.27 & 32.32 & 24.53 & 42.49 & 41.64 \\
    0.2 & 13732 & 32.74 & 57.44 & 32.56 & 25.64 & 42.39 & 41.26 \\
    0.3 & 14018 & 33.02 & 57.87 & 33.09 & 25.74 & 42.93 & 41.44 \\
    0.4 & 14018 & 33.02 & 57.87 & 33.09 & 25.74 & 42.93 & 41.44 \\
    \end{tabular}% <------ Don't forget this %
    }
    \caption{Impact on performance of the maximum crop size $\pi$ used the density crop labeling algorithm.}
    \label{table:max_clus_size_impact}
\end{table}

From these experiments on the hyperparameters of the crop labeling algorithm($\pi, \theta, N$), we can observe one thing in common; it is easy to tune their values. $N$ takes discrete values between(1, 3]. $\theta$ should be ideally small($< 0.2$) to have connections between scaled boxes. It is not sensible to use crops above 50\% of the size of the image. The maximum crop size should be at least 10\% of the image, else we won't get many crops. So $\pi$ should be something in between [0.1, 0.5).

%%%%%%%%%
\subsection{Impact of different backbones}
\begin{figure*}
\centering
\begin{tabular}{c @{\hspace{-0.3cm}} c  @{\hspace{-0.3cm}} c}
\includegraphics[width=5.6cm, height=4cm]{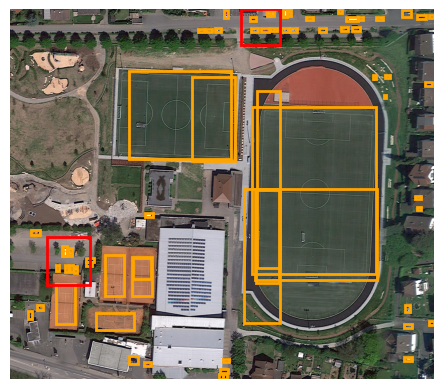} &
\includegraphics[width=5.6cm, height=4cm]{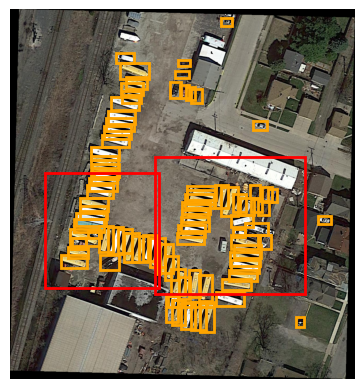} &
\includegraphics[width=5.6cm, height=4cm]{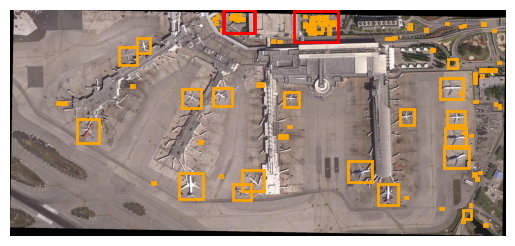} \\
\includegraphics[width=5.6cm, height=4cm]{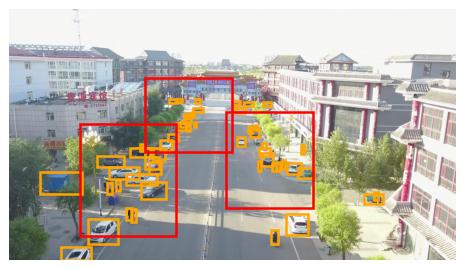} &
\includegraphics[width=5.6cm, height=4cm]{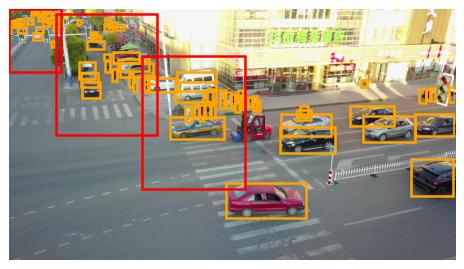} &
\includegraphics[width=5.6cm, height=4cm]{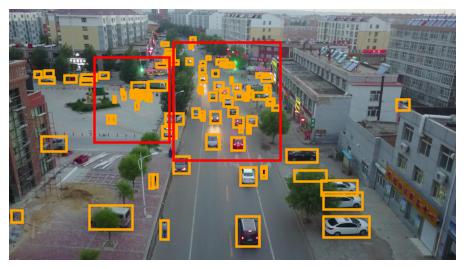} \\
 \end{tabular}
 \caption{Additional detection results: The first row shows detection on the DOTA images, the second row shows detection on the VisDrone images. Red boxes in each image shows the confident crops detected.}
 \label{fig:more_detections}
\end{figure*}

Table \ref{table:with_other_backbones} compares the result of our approach with ResNet-101 and ResNet-50\cite{resnet-He-2016} backbones. The trend is similar. For ResNet-101 also, the result without high-resolution features P2 is close to that of with P2, thus we are getting a significant boost even when high-resolution features are not used. This illustrates the advantages of using density crops over sparse convolutions on high resolution features for small object detection as proposed in \cite{querydet-Yang-2022}. Figure \ref{fig:more_detections} shows additional detection results on both VisDrone and DOTA datasets along with the high-quality density crops predicted by the detector.

\begin{table}
    \centering
    \resizebox{.48\textwidth}{!}{% <------ Don't forget this %
      \begin{tabular}{l||rrr|rrr|r}
    \textbf{Settings} & \textbf{AP} & \textbf{$\textrm{AP}_{50}$} & \textbf{$\textrm{AP}_{75}$} & \textbf{$\textrm{AP}_{s}$}  & \textbf{$\textrm{AP}_{m}$} & \textbf{$\textrm{AP}_{l}$} & FPS\\
    \hline
    \emph{Without P2} & & & & & & & \\
    Baseline R-50  & 29.48 & 51.68 & 29.55 & 22.33 & 38.66 & 39.30 & 26.31\\
    Baseline R-101 & 30.98 & 54.74 & 30.55 & 22.84 & 41.06 & 42.30 & 23.79\\    
    CZ Det. R-50 & 33.02 & 57.87 & 33.09 & 25.74 & 42.93 & 41.44 & 11.64\\
    CZ Det. R-101 & 33.91 & 59.07 & 33.87 & 26.14 & 44.64 & 45.03 & 10.01\\ 
    \hline
    \emph{With P2} & & & & & & & \\
    Baseline R-50 & 30.81 & 55.06 & 30.68 & 23.97 & 39.19 & 41.17 & 18.25\\
    Baseline R-101 & 31.41 & 55.47 & 31.29 & 23.97 & 40.56 & 43.83 & 16.48\\    
    CZ Det. R-50 & 33.22 & 58.30 & 33.16 & 26.06 & 42.58 & 43.36 & 8.44\\
    CZ Det. R-101 & 34.36 & 59.65 & 34.55 & 26.96 & 44.23 & 42.14 &  6.18\\
    \end{tabular}% <------ Don't forget this %
    }
    \caption{The performance of Baseline and CZ Detectors with the R-50 and R-101 backbones using Faster RCNN\cite{faster_rcnn-Ren-2015}.
    }
    \label{table:with_other_backbones}
\end{table}

\begin{figure}
     \centering
     \begin{subfigure}[b]{0.23\textwidth}
         \centering
         \includegraphics[width=\textwidth]{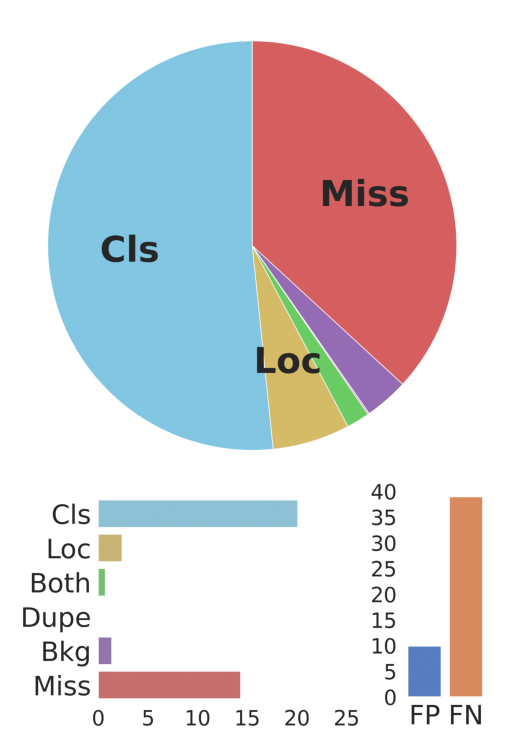}
         \caption{Baseline}
     \end{subfigure}
     \hfill
     \begin{subfigure}[b]{0.23\textwidth}
         \centering
         \includegraphics[width=\textwidth]{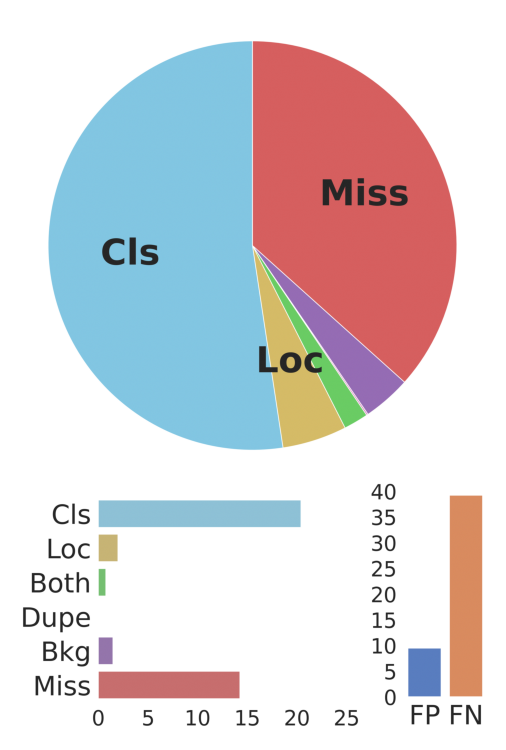}
         \caption{With density crops}
     \end{subfigure}
        \caption{Error analysis: Baseline vs With density crops. Error types are: \textbf{Cls}: localized correctly but classified incorrectly, \textbf{Loc}: classified correctly but localized incorrectly, \textbf{Both}: both cls and loc error, \textbf{Dupe}: duplicate detection error, \textbf{Bkg}: detected background as foreground, \textbf{Miss}: missed ground truth error.}
        \label{fig:tide_error_analysis}
\end{figure}

To understand whether our approach is creating any drastic changes in the learning dynamics of the baseline detector, we analyzed the error distribution of the detectors trained with our approach and the baseline training using TIDE\cite{tide-Bolya-eccv2020}. Figure \ref{fig:tide_error_analysis} shows the results. It is evident that learning a new class "density crop" with augmented crops and two-stage inference is not introducing any significant change in the behavior of the detector. The only change we observed is the reduction of the localization error, thus not altering any other aspects of the detector. Thus our method keeps the detector intact and only improves the performance of small object detection leveraging density crops.

